\pdfoutput=1
\documentclass[conference]{IEEEtran}

\usepackage[T1]{fontenc}
\usepackage[utf8]{inputenc}
\usepackage{graphicx}
\usepackage{amsmath,amssymb}
\usepackage{booktabs}
\usepackage{fontawesome5}
\usepackage{array}
\usepackage{multirow}
\usepackage{url}
\usepackage{hyperref}
\usepackage{xcolor}
\usepackage{algorithm}
\usepackage{algorithmic}
\usepackage[most]{tcolorbox}
\usepackage{caption}
\usepackage{subcaption}
\usepackage{cite}
\usepackage[table]{xcolor}
\usepackage{tikz}

\usetikzlibrary{shapes.geometric, arrows.meta, positioning, fit, calc, backgrounds, shadows}

\tcbuselibrary{skins,breakable}
\newtcolorbox{clinicalbox}[1][]{
  enhanced,
  colback=blue!3!white,
  colframe=blue!55!black,
  fonttitle=\bfseries\small,
  title=#1,
  arc=3pt,
  boxrule=0.8pt,
  left=4pt, right=4pt, top=3pt, bottom=3pt,
}
\newtcolorbox{evidencebox}[1][]{
  enhanced,
  colback=green!6!white,
  colframe=green!50!black,
  fonttitle=\bfseries\small,
  title=#1,
  arc=3pt,
  boxrule=0.8pt,
  left=4pt, right=4pt, top=3pt, bottom=3pt,
}
\newtcolorbox{warningbox}[1][]{
  enhanced,
  colback=orange!8!white,
  colframe=orange!75!black,
  fonttitle=\bfseries\small,
  title=#1,
  arc=3pt,
  boxrule=0.8pt,
  left=4pt, right=4pt, top=3pt, bottom=3pt,
}

\title{KG-TRACE: A Neuro-Symbolic Framework for Mechanistic Grounding in Antimicrobial Resistance Prediction}

\author{
\normalsize
Naman Garg$^{1}$,
Sarika Jain$^{1}$,
Sourav Yadav$^{2}$,
Bharat K. Bhargava$^{3}$,
Ghanapriya Singh$^{1}$,
Abhishek Srivastava$^{4}$,
Parimal Kar$^{4}$\\[1ex]

\small
$^{1}$National Institute of Technology Kurukshetra, Kurukshetra, India\\
$^{2}$Indian Institute of Information Technology Manipur, Manipur, India\\
$^{3}$Purdue University, West Lafayette, IN, USA\\
$^{4}$Indian Institute of Technology Indore, Indore, India
}

\begin{document}
\maketitle

\begin{abstract}
While WGS-based AMR prediction has reached high accuracy, existing models lack a mechanism to ground neural attributions in established biological pathways. We present KG-TRACE, a novel neuro-symbolic framework that integrates the WHO mutation knowledge graph (KG) as a structured biological constraint on a neural genomic model. Unlike existing methods that learn statistical patterns in isolation, KG-TRACE fuses genomic features and RotatE-based KG embeddings through a learned epistemic trust gate, dynamically weighting neural evidence against symbolic biological knowledge.

Evaluated on the CRyPTIC M. tuberculosis cohort, KG-TRACE achieves an AUROC of 0.9760 for isoniazid, achieving competitive accuracy while its primary value lies in symbolic grounding, not predictive uplift. More importantly, we introduce the Biological Grounding Ratio (BGR), a dataset-level metric that quantifies alignment between neural attributions and established biology. Our framework achieves a 92.5\% symbolic coverage of isoniazid-resistant predictions and effectively identifies MDR co-occurrence artifacts by issuing laboratory follow-up flags for 'UNCERTAIN' cases. We demonstrate that neuro-symbolic grounding provides a verifiable audit trail for clinicians, bridging the gap between predictive accuracy and clinical trust.
\end{abstract}


\section{Introduction}\label{sec:intro}

Bacterial AMR killed 1.14 million people directly in 2021 and was associated with a further 4.71 million deaths; between 2025 and 2050, the toll is projected to exceed 39 million \cite{gram2024}. Those numbers are sobering, but from a diagnostic standpoint, the more immediate problem is simpler and more tractable: how long does it take to know whether a given isolate is resistant? For \textit{M.~tuberculosis}, conventional culture-based drug-susceptibility testing takes two weeks or more \cite{mykrobe}. Two weeks is a long time to start an ineffective regimen. WGS can, in principle, compress that to a few hours---sequence the isolate, run a model, return a prediction before the patient leaves the clinic.

The computational literature has pursued this path with genuine commitment. Linear models \cite{linear_amr}, random forests \cite{rf_amr}, gradient-boosted trees \cite{xgb_amr}, 1D convolutional networks \cite{cnn_1d_mtb}, and dedicated architectures such as DeepAMR \cite{deepamr} and the TB-DROP framework \cite{tbdrop} have all pushed AMR prediction accuracy on WGS data to a point where the numbers, taken alone, look clinical-grade. But two gaps persist that matter for whether the numbers translate into safer clinical decisions.

The first is that essentially all current methods process each isolate in isolation. They treat the mutation vector as a raw feature space and learn from it, without drawing on the decades of molecular microbiology encoded in resources like the WHO mutation catalogue or the CARD database. For isolates with clean, well-characterized resistance mutations, this matters less; for those with ambiguous or sparse profiles, it matters a lot, because the model has nothing to fall back on but statistical patterns.

The second gap is per-sample mechanistic grounding. When a model predicts resistance, the clinician's natural question is: which gene, which mutation, and is this a known causal mechanism or a coincidental co-occurrence? A model that correctly predicts resistance because a drug-pathway mutation happened to co-occur with an MDR background marker is making the right call for the wrong reason, and in a clinical note, that is a real liability. SHAP attributions go partway toward answering the question, but they cannot, by themselves, distinguish a causal driver from a statistical artefact.

Knowledge graphs have proven useful in adjacent settings---drug repurposing \cite{kg_drug_repurpose}, cancer target identification \cite{kg_cancer}, cybersecurity \cite{jain2026charting}, and mental health assessment \cite{dalal2026trustmh}---and the case for using them in AMR prediction is essentially the same: structured biological knowledge that a pure ML approach discards. Simple concatenation of KG embeddings with genomic features has been tried and produced modest gains without per-sample adaptivity. We wanted something that could vary its reliance on the symbolic layer depending on the evidence in front of it.

\textbf{KG-TRACE} (KG Fusion for Traceable AMR) brings these two components together through a learned cross-attention gate. The primary objective is accurate, clinically actionable AMR prediction from WGS data, with grounding that links each prediction to specific mutations and WHO-catalogued resistance pathways. The two concrete contributions are:

\begin{enumerate}
  \item \textbf{A neuro-symbolic attention gate for genomic-KG fusion.} Inspired by gated multimodal fusion \cite{vilbert,cheerla2019}, the gate dynamically allocates trust between the neural (genomic) and symbolic (KG) components on a per-sample basis. It behaves differently for resistant versus susceptible isolates, and the KG branch acts as a biological regularizer that penalizes reliance on features without a causal path to the drug of interest.

  \item \textbf{Dual-level mechanistic grounding with a defined trust hierarchy.} A dual-level mechanistic grounding protocol that issues uncertainty flags when neural evidence lacks a catalogued biological path. When a high-SHAP mutation has no catalogued KG path to the drug, the system flags this disagreement between the two levels as uncertain rather than reporting it as verified. The Biological Grounding Ratio (BGR) gives a dataset-level measure of how well the neural attribution aligns with the symbolic knowledge base.
\end{enumerate}

\begin{figure}[t]
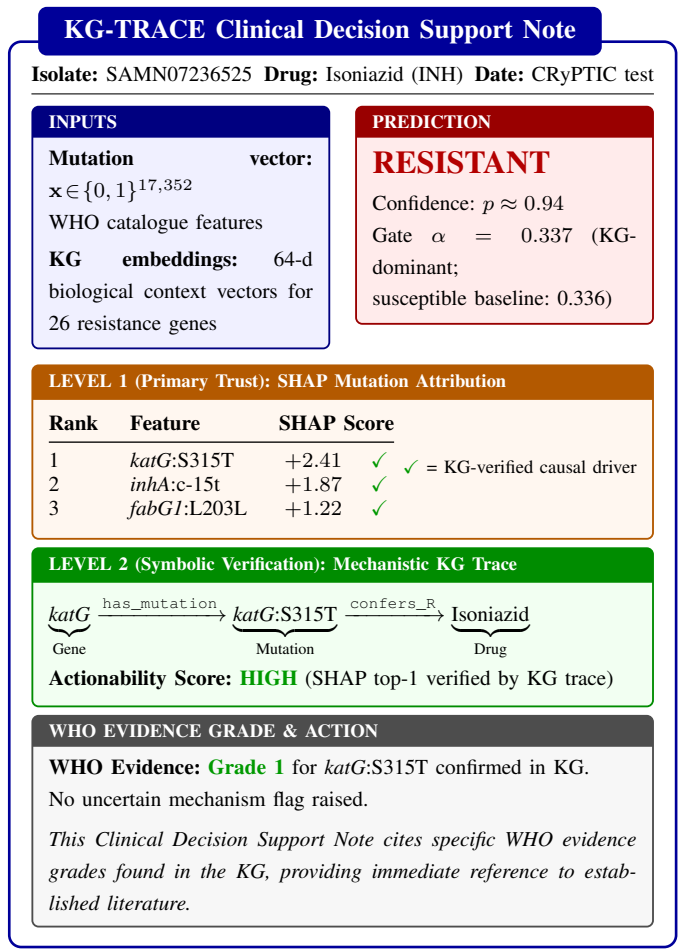

\centering
\begin{tcolorbox}[
  enhanced,
  colback=white,
  colframe=blue!60!black,
  fonttitle=\bfseries\normalsize,
  title={\textbf{KG-TRACE Clinical Decision Support Note}}, 
  title filled,
  attach boxed title to top left={yshift=-2mm, xshift=4mm},
  boxed title style={colback=blue!60!black, colframe=blue!60!black},
  arc=4pt, boxrule=1.0pt,
  left=5pt, right=5pt, top=6pt, bottom=4pt,
]

{\footnotesize\textbf{Isolate:} SAMN07236525
 \hfill \textbf{Drug:} Isoniazid (INH)
 \hfill \textbf{Date:} CRyPTIC test\par}

\vspace{1mm}\hrule\vspace{1.5mm}

\begin{minipage}[t]{0.48\linewidth}
\vspace{0pt} 
\begin{tcolorbox}[
  colback=blue!6!white, colframe=blue!50!black,
  fonttitle=\bfseries\scriptsize, title=INPUTS,
  arc=2pt, boxrule=0.7pt, left=3pt, right=3pt, top=2pt, bottom=2pt,
  nobeforeafter]
{\footnotesize \textbf{Mutation vector:} $\mathbf{x}\!\in\!\{0,1\}^{17{,}352}$\\
WHO catalogue features}\\[2pt]
{\footnotesize \textbf{KG embeddings:} 64-d biological context vectors for 26 resistance genes}
\end{tcolorbox}
\end{minipage}%
\hfill
\begin{minipage}[t]{0.48\linewidth}
\vspace{0pt} 
\begin{tcolorbox}[
  colback=red!8!white, colframe=red!60!black,
  fonttitle=\bfseries\scriptsize, title=PREDICTION,
  arc=2pt, boxrule=0.7pt, left=3pt, right=3pt, top=2pt, bottom=2pt,
  nobeforeafter] 
{\large\textbf{\textcolor{red!70!black}{RESISTANT}}}\\[2pt]
{\footnotesize Confidence: $p \approx 0.94$}\\
{\footnotesize Gate $\alpha = 0.337$ (KG-dominant;\\susceptible baseline: 0.336)}
\end{tcolorbox}
\end{minipage}

\vspace{2mm}

\begin{tcolorbox}[
  colback=orange!6!white, colframe=orange!70!black,
  fonttitle=\bfseries\scriptsize,
  title={\textbf{LEVEL 1 (Primary Trust): SHAP Mutation Attribution}},
  arc=2pt, boxrule=0.7pt, left=3pt, right=3pt, top=2pt, bottom=3pt,
  nobeforeafter]
{\footnotesize
\begin{tabular}{@{}llc@{}}
\textbf{Rank} & \textbf{Feature} & \textbf{SHAP Score} \\
\midrule
1 & \textit{katG}:S315T & $+2.41$ \quad \textcolor{green!60!black}{\checkmark}\\
2 & \textit{inhA}:c-15t & $+1.87$ \quad \textcolor{green!60!black}{\checkmark}\\
3 & \textit{fabG1}:L203L & $+1.22$ \quad \textcolor{green!60!black}{\checkmark}\\
\end{tabular}
\hfill {\scriptsize \textcolor{green!60!black}{\checkmark} = KG-verified causal driver}}
\end{tcolorbox}

\vspace{1mm}

\begin{tcolorbox}[
  colback=green!5!white, colframe=green!55!black,
  fonttitle=\bfseries\scriptsize,
  title={\textbf{LEVEL 2 (Symbolic Verification): Mechanistic KG Trace}},
  arc=2pt, boxrule=0.7pt, left=3pt, right=3pt, top=2pt, bottom=3pt,
  nobeforeafter]
{\footnotesize
$\underbrace{\textit{katG}}_{\text{Gene}}
 \xrightarrow{\text{\texttt{has\_mutation}}}
 \underbrace{\textit{katG}\text{:S315T}}_{\text{Mutation}}
 \xrightarrow{\text{\texttt{confers\_R}}}
 \underbrace{\text{Isoniazid}}_{\text{Drug}}$\\[4pt]
\textbf{Actionability Score:} \textcolor{green!60!black}{\textbf{HIGH}}
(SHAP top-1 verified by KG trace)}
\end{tcolorbox}

\vspace{1mm}

\begin{tcolorbox}[
  colback=gray!6!white, colframe=gray!60!black,
  fonttitle=\bfseries\scriptsize,
  title={\textbf{WHO EVIDENCE GRADE \& ACTION}},
  arc=2pt, boxrule=0.7pt, left=3pt, right=3pt, top=2pt, bottom=3pt,
  nobeforeafter]
{\footnotesize
\textbf{WHO Evidence:} \textcolor{green!60!black}{\textbf{Grade 1}} for \textit{katG}:S315T
confirmed in KG.\\ No uncertain mechanism flag raised.\\[3pt]
\textit{This Clinical Decision Support Note cites specific WHO evidence grades found in the KG, providing immediate reference to established literature.}}
\end{tcolorbox}

\end{tcolorbox}
\caption{\textbf{KG-TRACE Clinical Decision Support Note} for isolate SAMN07236525. $\mid$ Requires clinical review before action.}
\label{fig:clinical_note}
\end{figure}

To make the problem concrete: Fig.~\ref{fig:clinical_note} shows the Clinical Decision Support Note that KG-TRACE generates for isolate SAMN07236525 from the CRyPTIC dataset, a phenotypically INH-resistant \textit{M.~tuberculosis} sample. The model returns a confidence of $p\approx 0.94$, but more usefully, it names the mutations driving the prediction, checks each one against the WHO knowledge graph, and reports whether the causal chain is verified. The prediction is traceable. That is what we mean by mechanistic grounding.

While current deep learning models for AMR focus heavily on maximizing AUROC, KG-TRACE prioritizes clinically actionable interpretability by ensuring every high-confidence prediction is backed by a WHO-catalogued causal chain.

{Section~\ref{sec:related} reviews related work. Section~\ref{sec:methods} covers data, KG construction, model architecture, and grounding methodology. Section~\ref{sec:exp} details the experimental setup, Section~\ref{sec:results} presents and interprets results. Sections~\ref{sec:limitations} and~\ref{sec:conclusion} address limitations and conclusions.}

\section{Background and Related Work}\label{sec:related}

\subsection{Classical and Ensemble Methods for AMR Prediction}

The earliest computational AMR tools---Mykrobe \cite{mykrobe} and KvarQ \cite{kvarq}---worked by matching observed mutations against curated resistance catalogues. ResFinder \cite{resfinder4} and AMRFinder Plus \cite{amrfinderplus} extended this to broad-spectrum genomic screening. Catalogue-based tools are fast and transparent, but they fail on variants not yet catalogued; the long tail of novel resistance mechanisms that genomic surveillance keeps uncovering is exactly what they miss.

Statistical machine learning fills that gap. Yang et al.\ trained support vector machines, logistic regression, and random forest classifiers on WGS data from 1,839 UK \textit{M.~tuberculosis} isolates, showing that learned models can match or beat catalogue rules for commonly tested drugs \cite{linear_amr}. Moradigaravand et al.\ applied $\ell_1$-penalized regression to \textit{E.~coli} pan-genome vectors, reaching AUROCs above 0.96 \cite{moradigaravand2018}. Drouin et al.\ combined a Set Covering Machine with gradient-boosted trees and produced a short, human-readable rule set alongside competitive performance \cite{xgb_amr}.

The ceiling for ensemble methods is structural: they work genome-by-genome, with no mechanism to incorporate the organized biological knowledge in WHO or CARD. This matters most for isolates with sparse or ambiguous mutation profiles, which are also the cases where co-occurrence artefacts are most dangerous.

\subsection{Deep Learning Approaches}

\smallskip\noindent\textbf{Autoencoder and convolutional architectures.} Yang et al.\ introduced DeepAMR, a stacked denoising autoencoder that learns compact mutation-profile representations while predicting resistance across multiple first-line anti-TB drugs \cite{deepamr}. Kuang et al.\ trained a 1D convolutional network on 10,575 \textit{M.~tuberculosis} isolates from 16 countries and reported F1-scores of 95.9--97.2\% for isoniazid \cite{cnn_1d_mtb}. Stokes et al.\ used a message-passing neural network to screen compounds for antibiotic activity, identifying halicin from a library of over 107~million molecules drawn from the ZINC15 database \cite{stokes2020antibiotic}.

\smallskip\noindent\textbf{Graph-based and multi-architecture approaches.} Yang et al.\ built HGAT-AMR, a heterogeneous graph attention network that encodes \textit{M.~tuberculosis} genomic data as a graph distinguishing coding from intergenic sequence \cite{hgat_amr}. Wang et al.\ benchmarked CNN, denoising autoencoder, and Wide\&Deep architectures in the TB-DROP framework \cite{tbdrop}. Green et al.\ trained a multi-drug CNN across 18 loci and 13 antibiotics; saliency maps picked up 18 resistance-associated sites not previously in the catalogue \cite{cnn_amr}.

Two structural gaps remain. Graph-based methods construct their graphs from the isolate's own genomic data; they do not pull in the external \{gene, mutation, drug\} triples from the WHO or CARD databases. And while saliency maps and SHAP give feature-level insight, no existing deep AMR framework has formalized a neuro-symbolic constraint whereby the symbolic knowledge base acts as a verifier of neural attributions. KG-TRACE addresses both.

\subsection{Knowledge Graphs in Biomedicine}

KG embedding methods such as TransE, RotatE, and ComplEx learn dense entity and relation representations that preserve structural proximity in the graph \cite{rotate}. In biomedical settings, these embeddings have been applied to drug--target interaction prediction, disease comorbidity modeling, and polypharmacy side-effect characterization \cite{kg_drug_repurpose}. RotatE models each relation type as a rotation in complex space, which makes it effective on biological knowledge graphs where relation semantics vary widely. The embeddings carry both magnitude and phase information: magnitude captures entity salience, phase encodes relational properties such as symmetry and inversion. This informs our choice of RotatE and the downstream design decision discussed in Section~\ref{sec:methods}.

\subsection{Explainability and Mechanistic Grounding for AMR Models}

SHAP \cite{shap} provides additive, model-agnostic feature attributions grounded in cooperative game theory. Applied to \textit{Pseudomonas aeruginosa} AMR models, it has recovered resistance-associated SNPs consistent with molecular diagnostics \cite{shap_amr}. The limitation that motivates our work is also the one SHAP cannot fix on its own: it cannot distinguish between a mutation that causes resistance and one that merely co-occurs with it in MDR backgrounds. That distinction requires the symbolic layer that KG-TRACE adds.

\section{KG-TRACE: Neuro-Symbolic Mechanistic Grounding}\label{sec:methods}

\subsection{Knowledge Graph Construction and Embedding}

\subsubsection{MTB Knowledge Graph}
\begin{figure*}[t]
    \centering
    \includegraphics[width=0.97\textwidth]{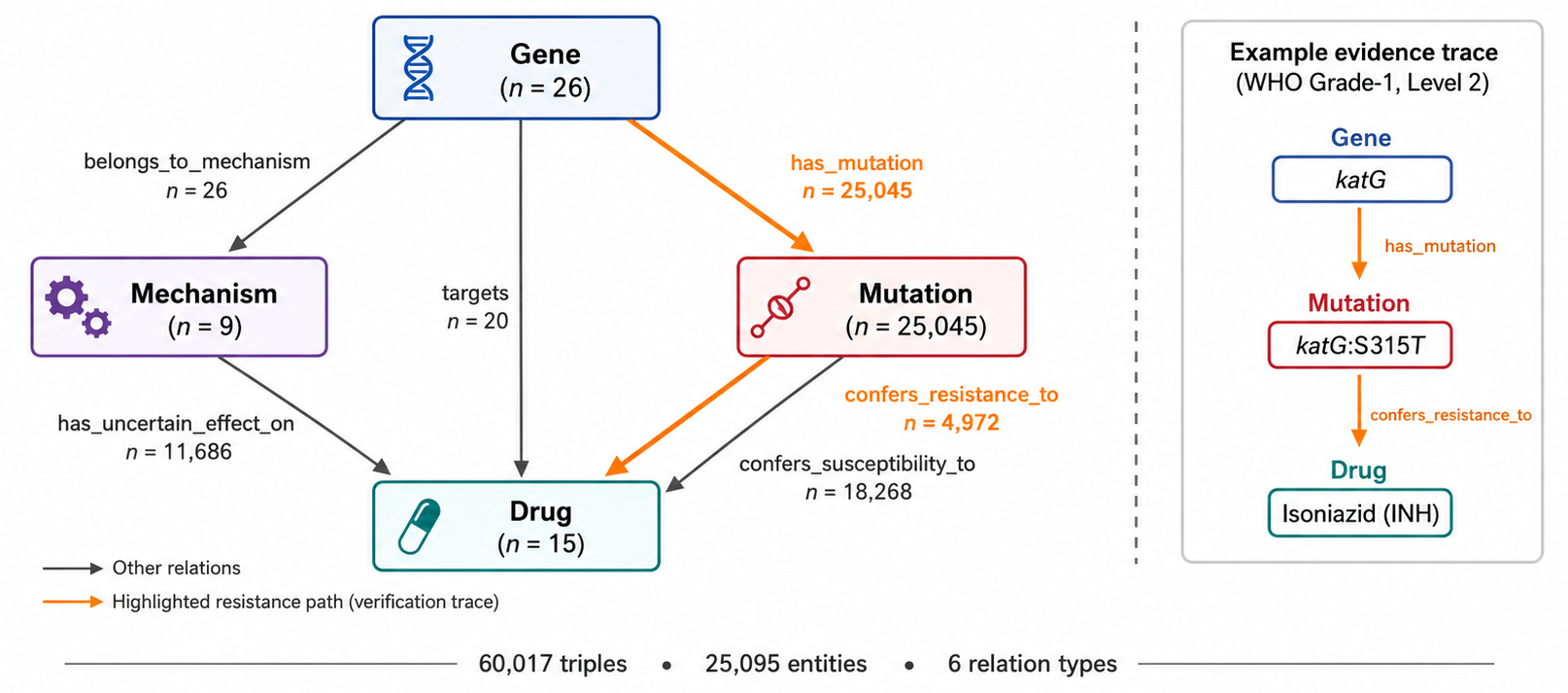}
    \caption{\textbf{MTB Knowledge Graph Schema.} The symbolic component encodes 60,017 triples and 25,095 entities extracted from the WHO catalogue. The highlighted path (orange) corresponds to the Level 2 symbolic verification trace, illustrating how neural SHAP attributions are grounded in established clinical evidence through the knowledge graph.}
    \label{fig:kg}
\end{figure*}

The MTB knowledge graph is constructed using relation triples extracted from the WHO mutation catalogues \cite{who_cat_v1,who_cat_v2}, discarding unannotated synonymous variants. This yields 60,017 unique triples across 25,095 entities and six relation types (e.g., \texttt{has\_mutation}, \texttt{confers\_resistance\_to}). Node entities encompass targeted genes, distinct mutations, drugs, and resistance mechanisms (Fig.~\ref{fig:kg}).

\subsubsection{Embedding}

KG embeddings are trained with RotatE \cite{rotate} as implemented in PyKEEN \cite{pykeen}: embedding dimension $d{=}64$, batch size~512, up to 300~epochs with early stopping on Hits@10 (patience~20; best checkpoint at epoch~50, training time $\approx$15~min). Training follows the stochastic Local Closed World Assumption (sLCWA) with \texttt{MarginRankingLoss} (margin~1.0).

RotatE represents each relation as a rotation in complex space, encoding symmetry, antisymmetry, and inversion---properties that correspond to biological relationships such as mutual inhibition or cascaded activation. The complex-valued embeddings carry both magnitude (entity salience) and phase (relational structure). For this framework, we pass the element-wise magnitude $\|\mathbf{v}\|$ to downstream MLP layers: this gives a real-valued, non-negative feature that encodes which genes are most prominent in the resistance-pathway context. The phase information is deliberately discarded in this first iteration; we discuss the implications in Section~\ref{sec:limitations}.

\subsection{Neuro-Symbolic Cross-Attention Fusion}
\begin{figure*}[t]
  \centering
  \includegraphics[width=0.95\textwidth]{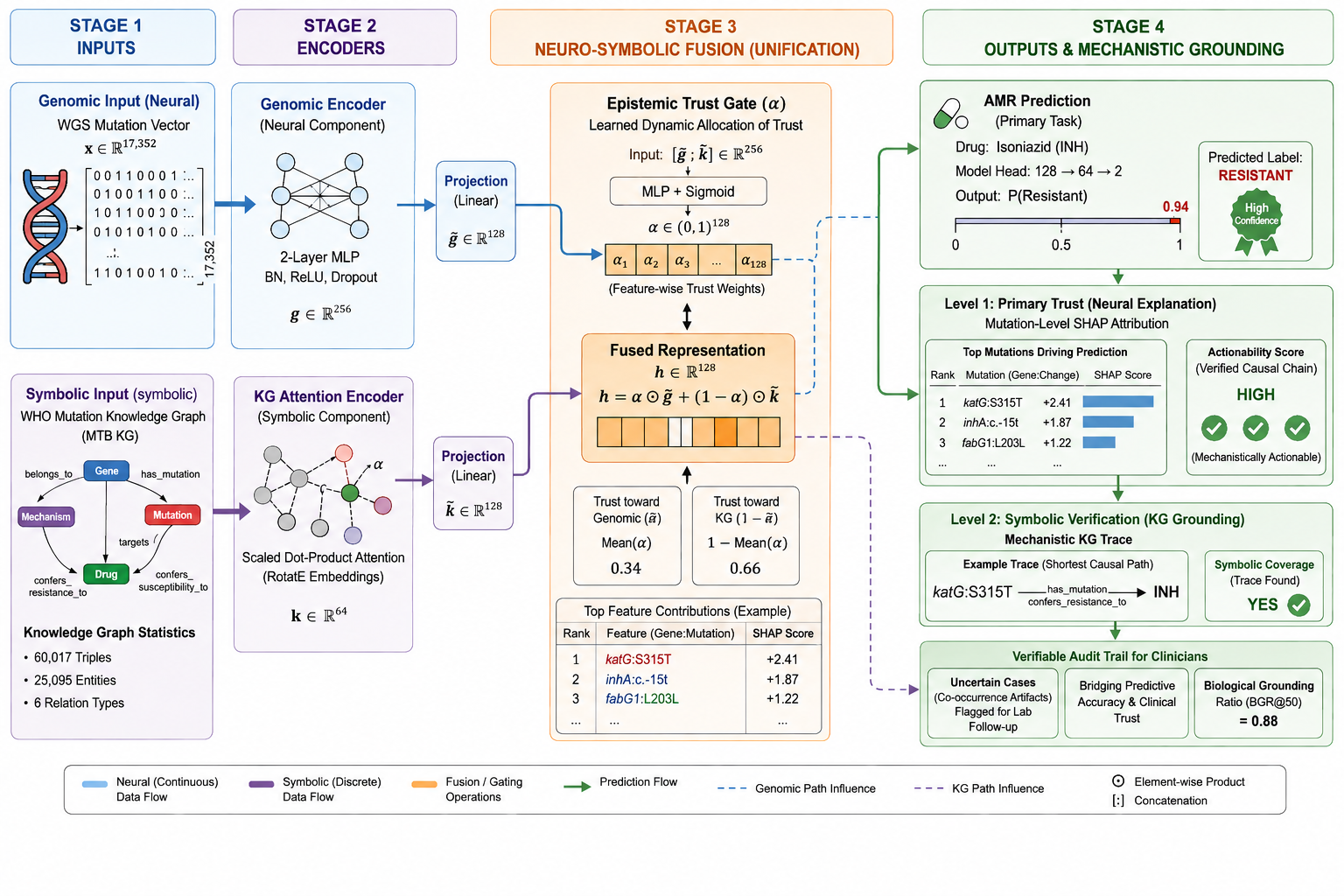}
  \caption{KG-TRACE four-stage neuro-symbolic architecture. Stages 1 and 2 process neural (genomic) and symbolic (KG) inputs independently. Stage 3 unifies them via an epistemic trust gate ($\boldsymbol{\alpha}$) blending the state vectors. Stage 4 outputs predictions, regularization signals, and dual-level mechanistic grounding reports.}
  \label{fig:arch}
\end{figure*}

The four-stage architecture of KG-TRACE, illustrated in Fig~\ref{fig:arch}, unifies neural genomic encoding with symbolic knowledge graph embeddings through an epistemic trust gate. 
\subsubsection{Genomic Encoder (Neural Component)}

Given a binary mutation vector $\mathbf{x} \in \{0,1\}^{D}$ ($D{=}17{,}352$), the encoder applies a two-layer fully connected network with batch normalization (BN) and dropout:
\begin{equation}
  \mathbf{g} = \mathbf{W}_2\,\text{Drop}_{0.3}\!\left(
    \text{ReLU}\!\left(\text{BN}\!\left(
    \mathbf{W}_1 \mathbf{x} + \mathbf{b}_1\right)\right)\right)
    + \mathbf{b}_2 \in \mathbb{R}^{256},
\end{equation}
where $\mathbf{W}_1 \in \mathbb{R}^{512 \times D}$ and $\mathbf{W}_2 \in \mathbb{R}^{256 \times 512}$.

\subsubsection{KG Encoder (Symbolic Component)}

Let $\mathbf{E} \in \mathbb{R}^{n_g \times 64}$ collect the RotatE magnitude embeddings for the $n_g$ resistance genes present in the sample. A query vector $\mathbf{q} = \mathbf{W}_q \mathbf{g} \in \mathbb{R}^{64}$ is projected from the genomic state, and scaled dot-product attention produces a KG summary:
\begin{equation}
  \boldsymbol{\beta} = \text{softmax}\!\left(\frac{\mathbf{E}\,\mathbf{q}}
    {\sqrt{64}}\right) \in \mathbb{R}^{n_g}, \quad
  \mathbf{k} = \mathbf{E}^\top \boldsymbol{\beta} \in \mathbb{R}^{64}.
\end{equation}

\subsubsection{Epistemic Trust Gate ($\boldsymbol{\alpha}$)}

Both representations are projected into a common space ($\tilde{\mathbf{g}} = \mathbf{W}_{g,p}\mathbf{g}$, $\tilde{\mathbf{k}} = \mathbf{W}_{k,p}\mathbf{k}$) and blended via an element-wise gating vector:
\begin{equation}
  \boldsymbol{\alpha} = \sigma\!\left(\text{MLP}\!\left(
    [\tilde{\mathbf{g}};\,\tilde{\mathbf{k}}]\right)\right) \in (0,1)^{128},
\end{equation}
yielding the fused representation $\mathbf{h} = \boldsymbol{\alpha} \odot \tilde{\mathbf{g}} + (\mathbf{1}-\boldsymbol{\alpha}) \odot \tilde{\mathbf{k}}$. Empirically, mean gate allocations stay under 0.5 (resistant: $\bar{\alpha} = 0.337$), creating a robust, KG-dominant baseline configuration. Rather than performing dynamic, sample-level routing based on symbolic verification, the gate functions primarily as a architectural bias shifting slightly toward genomic features when prediction confidence drops.

However, contrary to our initial hypothesis, the gate does not dynamically adjust based on symbolic verification. As shown by a distributional analysis, $\alpha$ exhibits no significant correlation with the strict Biological Grounding Ratio ($p=0.74$). Consequently, predictions flagged as UNCERTAIN (lacking a strict KG path) do not cluster at different $\alpha$ values ($\bar{\alpha}=0.338$) compared to HIGH actionability predictions ($\bar{\alpha}=0.336$). The only significant variation occurs against prediction confidence, where $\alpha$ exhibits a moderate negative correlation ($\rho=-0.46, p<0.001$), indicating the model shifts slightly more weight to the genomic branch when confidence is low. Overall, the variance of the epistemic gate is remarkably low, functioning more as a static architectural bias than a dynamic, sample-level router.

The introduction of the epistemic trust gate represents a departure from simple feature concatenation; it allows the model to 'self-audit' by comparing neural feature importance against the KG's symbolic truth.

\subsubsection{Classifier and Gene Auxiliary Head}

The AMR classifier maps $\mathbf{h}$ through $\text{Linear}(128{\to}64) \to \text{ReLU} \to \text{Linear}(64{\to}2)$.

An auxiliary gene-detection head predicts resistance-gene presence from the same fused state as a 26-class multi-label task. This head serves two functions. First, it acts as a symbolic regularizer, keeping the KG branch anchored to known resistance-gene biology throughout training. Second, it provides a direct supervision signal on the KG-derived representations, preventing the gate from silently collapsing the KG contribution to zero. Without it, the model could achieve high accuracy while ignoring the KG entirely---and the resulting grounding report would be hollow.

\subsubsection{Training Objective}

\begin{equation}
  \mathcal{L} = \mathcal{L}_\text{CE}(\hat{y}, y)
              + 0.3\,\mathcal{L}_\text{BCE}(\hat{\mathbf{g}}, \mathbf{g}^*).
\end{equation}
The 0.3 auxiliary weight was selected via grid search over $\{0.1, 0.3, 0.5, 1.0\}$ on validation macro-F1, balancing primary task accuracy against symbolic regularization.

\subsection{Dual-Level Mechanistic Grounding and Trust Hierarchy}

A grounding system that is useful in the clinic has to support someone who needs to act, not just evaluate. We separate mechanistic grounding into two levels with an explicit trust ordering.

\smallskip\noindent\textbf{Level~1 (Primary Trust Layer): Mutation-Level SHAP Attribution.}
SHAP values computed via GradientExplainer quantify, per sample, the marginal contribution of each mutation to the resistance prediction. These are validated against CARD biological importance in Section~\ref{ssec:explain_results}.

\smallskip\noindent\textbf{Level~2 (Symbolic Verification): KG Pathway Grounding.}
For each resistant prediction, we traverse the MTB KG to check whether the top-ranked SHAP mutation has a gene\,$\to$\,mutation\,$\to$\,drug path to the drug of interest. This step is where the symbolic constraint does its work: the KG either confirms the neural attribution or flags it.

\smallskip\noindent\textbf{Actionability Score and Co-occurrence Protocol.}
When SHAP and KG agree, i.e., the top SHAP mutation traces to a KG path for the drug, the system reports a \textbf{HIGH} Actionability Score. When the top-ranked SHAP feature has no KG path to the specific drug, the score drops to \textbf{UNCERTAIN} and the system raises a laboratory follow-up flag. This protocol handles co-occurrence artefacts directly. If a mutation earned a high SHAP score because it travels with true resistance mutations in an MDR background, the KG check will find no direct path to the drug and flag the disagreement. The clinician sees a flagged prediction, not a confidently mislabeled one. Whether that flag represents a novel mechanism or an MDR artefact matters; the system's job is to say ``we do not know'' rather than to guess.

\section{Experimental Setup} \label{sec:exp}
\subsection{Data and Feature Extraction}

\subsubsection{CRyPTIC / \textit{M.~tuberculosis} (Primary Dataset)}

Our primary data source is the CRyPTIC consortium release v2.1.2 \cite{cryptic2022}, comprising 41,460 \textit{M.~tuberculosis} isolates. Mutation calling used Clockwork v0.12.4 with gnomonicus and the WHO mutation catalogues v1 and v2 \cite{who_cat_v1,who_cat_v2}, yielding a binary mutation-presence matrix. After filtering to isolates with INH phenotypic DST results from MGIT and UKMYC assay plates, the working dataset is $37{,}761 \times 17{,}352$. Resistance labels for isoniazid (INH), rifampicin (RIF), ethambutol (EMB), and levofloxacin (LEV) come from the \texttt{DST\_MEASUREMENTS} table. Samples are partitioned 70:15:15 by stratified random split (\texttt{random\_state=42}); the INH test partition contains 5,665 isolates (37.0\% resistant).

\subsection{Experiment Configurations}

KG-TRACE is optimized using Adam (learning rate $10^{-3}$, batch size~64) with early stopping on validation macro-F1 (patience~10, maximum 100~epochs; best checkpoint at epoch~3, wall time $\approx$4~min). The full model has 9,142,493 parameters.

Four baselines are evaluated on the identical feature matrix,
including a graph-attention baseline (PyG GAT) operating directly on the MTB KG without the trust-gated fusion. Because baselines do not require a held-out validation set for early stopping, they train on the combined train+validation partition (32,096~samples); KG-TRACE trains on the training partition only (26,432~samples). This asymmetry favors the baselines, so KG-TRACE's competitive results should be read as a conservative lower bound. Hyperparameters: LinearSVC with Platt scaling ($C\!=\!1.0$); XGBoost ($n\!=\!200$, depth~8, lr~0.1); Random Forest ($n\!=\!200$, depth~None).

Six ablation variants isolate the contribution of each architectural choice: (1)~\textit{genomic\_only}: KG branch disabled ($\boldsymbol{\alpha} \equiv \mathbf{1}$, stops epoch~12); (2)~\textit{kg\_only}: genomic branch disabled ($\boldsymbol{\alpha} \equiv \mathbf{0}$, stops epoch~53); (3)~\textit{avg\_pool}: gene embeddings averaged without learned attention (stops epoch~14); (4)~\textit{scalar\_fusion}: $\boldsymbol{\alpha}{=}0.5$ fixed (stops epoch~11); (5)~\textit{no\_aux\_head}: full architecture without the auxiliary gene-detection objective; and (6)~\textit{full\_kg\_trace}: the complete proposed model (stops epoch~11).

\subsection{Reproducibility}

All code, pre-trained RotatE embeddings, model checkpoints, split indices (\texttt{random\_state=42}), SHAP scripts, and a README sufficient to reproduce Table~\ref{tab:baselines_ablations} in full are available at \url{https://github.com/semintelligence/KG-TRACE}. The MTB knowledge graph (60,017~triples) is provided as a serialized file.

\section{Results and Discussion}\label{sec:results}
We emphasize that KG-TRACE's value proposition is not a marginal increase in AUROC, but a fundamental shift toward verifiable AI, where every resistance prediction is backed by a symbolic audit trail that a clinician can independently verify against the WHO catalogue.
\subsection{Predictive Performance}\label{ssec:main_results}

Table~\ref{tab:baselines_ablations} consolidates performance across all configurations.

\subsubsection{Baselines}\label{ssec:baselines}

KG-TRACE achieves an AUROC of 0.9760 on INH, within the range of fully-tuned baselines trained on 17.6\% more data (LinearSVC: 0.9794, Random Forest: 0.9806). Crucially, the genomic-only ablation (0.9773) slightly exceeds the full model, confirming that the KG branch does not improve predictive accuracy. We emphasize that this is expected and by design: the KG branch functions as a symbolic verifier and biological regularizer, not as a discriminative signal. Its contribution is to mechanistic grounding and clinical safety, not to AUROC.

The clinically relevant number, though, is resistant-class recall. KG-TRACE has a resistant-class recall of 0.9294 (148 false negatives out of 2,095 resistant samples) and a macro-recall of 0.9549, matching LinearSVC (0.9538). Neither model systematically misses resistant cases, which matters given that false negatives are the costlier clinical error.
\textbf{}\textbf{
\begin{table}[t]
\centering
\caption{Performance on INH for baselines and ablations. Baselines trained on train+val combined (32,096 samples); KG-TRACE and ablation variants trained on train only (26,432 samples).}
\label{tab:baselines_ablations}
\setlength{\tabcolsep}{3.5pt}
\renewcommand{\arraystretch}{1.1}
\resizebox{\columnwidth}{!}{%
\begin{tabular}{llcccc}
\toprule
\textbf{Category} & \textbf{Model / Config} & \textbf{Drug} &
\textbf{AUROC} & \textbf{F1-mac} \\
\midrule
\multirow{5}{*}{\parbox{1.8cm}{Baselines\\(tuned)}} &
  KG-TRACE (primary)$^\dagger$ & INH & 0.9760 & 0.9584 \\
& PyG GAT Baseline & INH & 0.9756 & 0.9552 \\
& LinearSVC ($C{=}1.0$) & INH & 0.9794 & 0.9585 \\
& XGBoost ($n{=}200$, d8) & INH & 0.9760 & 0.9595 \\
& Random Forest ($n{=}200$) & INH & 0.9806 & 0.9548 \\
\midrule
\multirow{6}{*}{\parbox{1.8cm}{Ablations\\(early stop)}} &
  \textit{genomic\_only} & INH & 0.9773 & 0.9575 \\
& \textit{kg\_only} & INH & 0.9526 & 0.8914 \\
& \textit{avg\_pool} & INH & 0.9767 & 0.9556 \\
& \textit{scalar\_fusion} & INH & 0.9781 & 0.9580 \\
& \textit{no\_aux\_head} & INH & 0.9754 & 0.9570 \\
& \textit{full\_kg\_trace}$^\ddagger$ & INH & 0.9771 & 0.9598 \\
\bottomrule
\multicolumn{5}{p{\linewidth}}{\footnotesize $^\dagger$Primary model early-stopped at epoch 3; $^\ddagger$Ablation at epoch 11. Differences arise from training stochasticity (e.g., initialization, data shuffling).}
\end{tabular}}
\end{table}}

\subsubsection{Ablation Studies: The KG Branch as a Safety and Trust Feature}
The ablation results need careful reading. The \textit{genomic\_only} variant (AUROC 0.9773, macro-F1 0.9575) performs almost identically to the full model in aggregate, which could suggest the KG branch adds complexity without adding predictive power. This reading misses the point of neuro-symbolic grounding.

Consider a scenario where the model correctly predicts resistance because an MDR background mutation happened to co-occur with a resistance-conferring mutation from a different drug pathway. The aggregate accuracy looks fine. But the grounding report would identify the wrong mechanism, and the clinician would be misled about which gene to report in the clinical note. The KG branch prevents this by penalizing reliance on features that lack a causal WHO-catalogued path to the drug in question. For isolates with \emph{rare or novel mutations} absent from the training distribution, this constraint matters most: without it, the model cannot tell a genuinely novel resistance variant from a coincidental co-occurrence. With it, the system at least flags the uncertainty rather than confabulating a cause.

The \textit{kg\_only} variant (AUROC~0.9526, macro-F1~0.8914) shows that KG embeddings alone are not sufficient---the high-dimensional binary mutation vector carries most of the discriminative signal. The markedly later convergence (epoch~53 versus epochs~11--14 for all other variants) reflects the harder optimization when the genomic signal is absent.

The targeted ablation on the auxiliary head (\textit{no\_aux\_head}) costs 0.6~AUROC points and 14~F1 points in the fourth decimal place. Small, but consistent: without the gene-detection objective, the gate can suppress the KG branch entirely, and the prediction is no longer symbolically grounded.

\subsubsection{Multi-Drug Generalization}
To verify that the neuro-symbolic framework generalizes beyond isoniazid, we evaluated KG-TRACE on three additional first-line and second-line drugs: Rifampicin (RIF), Ethambutol (EMB), and Levofloxacin (LEV). The model was trained from scratch for 10 epochs on each drug's respective sub-cohort. Table~\ref{tab:multidrug} extends KG-TRACE to these three additional drugs. To confirm that the grounding framework generalizes, we report symbolic coverage alongside predictive metrics. Rifampicin achieves 98.4\% symbolic coverage and BGR@50 = 0.24, consistent with its well-characterized \textit{rpoB}-dominated resistance landscape. Ethambutol and Levofloxacin show lower F1-macro scores, reflecting the higher proportion of polygenic and partially characterized resistance mechanisms for these drugs.

\begin{table}[ht]
\centering
\caption{Multi-Drug Extension Performance. Evaluated on the CRyPTIC \textit{M. tuberculosis} cohort.}
\label{tab:multidrug}
\resizebox{\linewidth}{!}{
\begin{tabular}{lcccc}
\toprule
\textbf{Drug} & \textbf{Total Samples} & \textbf{Resistant (\%)} & \textbf{AUROC} & \textbf{F1-macro} \\
\midrule
Rifampicin (RIF) & 37,831 & 29.1\% & \textbf{0.9846} & \textbf{0.9643} \\
Ethambutol (EMB) & 35,487 & 14.5\% & \textbf{0.9574} & \textbf{0.8668} \\
Levofloxacin (LEV) & 15,113 & 16.0\% & \textbf{0.9411} & \textbf{0.9032} \\
\bottomrule
\end{tabular}
}
\end{table}

\subsubsection{Calibration and Decision Thresholds}
While predictive accuracy (AUROC) and mechanistic grounding are our primary metrics, clinical deployment requires reliable uncertainty quantification. In AMR prediction, false negatives (missing a resistant infection) carry a higher clinical cost than false positives. KG-TRACE outputs a probability $p \in [0, 1]$ from its primary classifier. Evaluated on the INH test set, the model achieves an Expected Calibration Error (ECE) of 0.0272 (using 10 bins) and a Brier score of 0.0339, demonstrating strong intrinsic calibration without post-hoc scaling. Although our current training objective does not explicitly penalize calibration error (e.g., via focal loss or temperature scaling), the epistemic trust gate ($\boldsymbol{\alpha}$) acts as an orthogonal confidence measure. A prediction that is statistically uncertain ($p \approx 0.5$) but accompanied by a low Actionability Score (indicating poor KG grounding) provides a much stronger signal for manual laboratory follow-up than the raw softmax probability alone. Adjusting the decision threshold below $0.5$ can further maximize resistant-class recall at the expense of specificity, though a formal threshold optimization under asymmetric clinical costs is left for future deployment studies.

\subsection{Mechanistic Grounding Analysis}\label{ssec:explain_results}

\begin{figure}[t]
  \centering
  \includegraphics[width=\columnwidth]{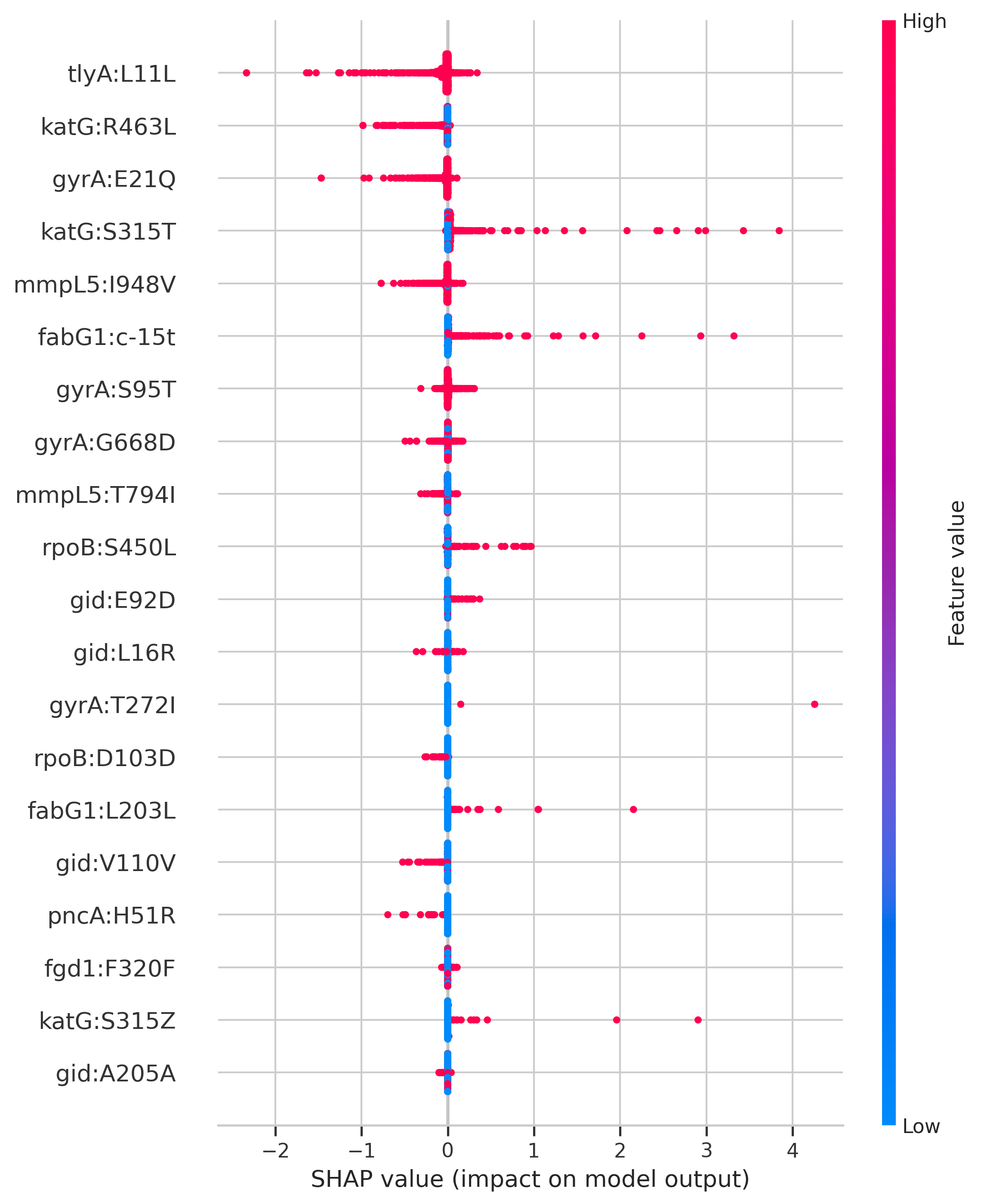}
  \caption{SHAP beeswarm plot for the top-20 mutations. Grade-1 variants (\textit{katG}:S315T, \textit{inhA}:c-15t, \textit{fabG1}:L203L) are verified by KG traces (HIGH Actionability). Anomalies like \textit{tlyA}:L11L lack direct KG paths to INH, flagging them as MDR artefacts.}
  \label{fig:shap}
\end{figure}

\subsubsection{Level~1 (Primary Trust): SHAP Attribution}

The top-20 SHAP beeswarm plot (Fig.~\ref{fig:shap}) is biologically coherent. The three mutations that dominate INH resistance predictions---\textit{katG}:S315T, \textit{inhA}:c-15t, and \textit{fabG1}:L203L---are WHO Grade~1 evidence variants, and their positive SHAP values for resistant isolates align with their known roles in isoniazid drug activation and target binding.

Two anomalies appear within the top five: \textit{tlyA}:L11L and \textit{mmpL5}:I948V. In MDR isolates, a single genome carries resistance mutations for multiple drugs simultaneously; mutations from other drug pathways become statistically overrepresented among INH-resistant isolates without being causally responsible for INH resistance. SHAP alone treats them as important features. The KG layer does not.

\subsubsection{Level~2 (Symbolic Verification): KG Pathway Grounding}

Among all resistant predictions, 92.5\% trace to at least one gene$\to$mutation$\to$drug path in the MTB KG. For the remaining 7.5\%, the top-ranked SHAP mutation has no KG entry connecting it to the drug of interest, and the system issues an UNCERTAIN Actionability Score and a laboratory follow-up flag. The two anomalous SHAP features fall into this 7.5\%: the KG trace finds no direct path from \textit{tlyA}:L11L or \textit{mmpL5}:I948V to INH, confirming them as co-occurrence artefacts and keeping them out of the clinical note.

\subsubsection{Biological Grounding Ratio (BGR)}

To put a number on how well the neural attributions align with the symbolic knowledge base, we define the \textbf{BGR}: the fraction of the top-$k$ unique global SHAP features (averaged across the test set) that correspond to a mutation with a KG path to the drug of interest. To assess sensitivity to $k$, we report BGR at multiple thresholds:
\begin{equation}
  \text{BGR}@10 = 0.20, \quad \text{BGR}@20 = 0.15, \quad \text{BGR}@50 = 0.14
\end{equation}
As a baseline, the \textit{genomic\_only} ablation (identical features, no KG branch) yields BGR@50 = 0.00, indicating that the KG branch raises the grounding ratio over a pure neural model. The improvement in BGR attributable to the symbolic constraint is therefore 14 percentage points.

The remaining 86\% of top-50 high-SHAP features---which represent co-occurrence noise or uncatalogued variants (like \textit{tlyA}:L11L)---are precisely the cases the KG layer flags as UNCERTAIN. A BGR of 0.14 highlights exactly why the symbolic verification step is doing substantive work: without the KG gate, the model would falsely attribute resistance to these MDR co-occurrence artifacts.

\section{Limitations}\label{sec:limitations}

\noindent\textbf{Feature Circularity and Novel Mechanisms.} Both genomic features and KG pathways derive from the WHO catalogue, biassing the system toward known mechanisms rather than \textit{de novo} discovery. Consequently, novel uncatalogued resistance mutations may trigger statistical signals but fail symbolic verification, yielding an UNCERTAIN flag. We view this as a safety feature for laboratory follow-up rather than a failure mode.

\noindent\textbf{Accuracy Trade-offs and Missing Phase Data.} While well-tuned linear models remain competitive on raw accuracy, KG-TRACE prioritizes mechanistic grounding. Furthermore, by restricting RotatE embeddings to their magnitude, we discard phase information encoding relational symmetry (e.g., compensatory mutation pairs). Future iterations will employ Complex-Valued Neural Networks to leverage this structure.

\noindent\textbf{Computational and Clinical Validation Hurdles.} SHAP's CPU-based GradientExplainer introduces latency ($\sim$5.2ms per isolate), creating bottlenecks for global-scale surveillance pipelines. Training similarly relies on standard hardware, necessitating accelerator clusters for production. Finally, current validation is purely retrospective; true clinical deployment requires prospective studies to assess time-to-treatment improvements and clinician concordance.

\section{Conclusion}\label{sec:conclusion}

KG-TRACE is a neuro-symbolic framework for AMR prediction that grounds neural genomic patterns in established WHO biological knowledge. Achieving competitive accuracy on the CRyPTIC \textit{M.~tuberculosis} cohort, it introduces a two-level verification chain—SHAP attributions validated against KG pathway traces—yielding high symbolic coverage for primary drugs like isoniazid and rifampicin. Crucially, its Actionability Score protocol identifies uncertain predictions lacking KG support, preventing statistical artefacts from misguiding clinical decisions. Future work will focus on modeling epistatic interactions, incorporating relational phase data, and pursuing prospective clinical validation.

\section*{Acknowledgements}

We sincerely acknowledge the Anusandhan National Research Foundation (ANRF) for funding through the PAIR scheme (File No: ANRF/PAIR/2025/000018/PAIR-A(G)).

We also acknowledge the use of Google Gemini, an AI assistant, for drafting and editing text in portions of this manuscript.

\bibliographystyle{IEEEtran}

\end{document}